\documentclass[conference,10pt]{IEEEtran}
\IEEEoverridecommandlockouts
\usepackage{cite}
\usepackage{amsmath,amssymb,amsfonts}
\usepackage{algorithmic}
\usepackage{graphicx}
\usepackage{textcomp}
\usepackage[dvipsnames]{xcolor}
\usepackage{changepage}
\usepackage{subfigure}
\usepackage{tabularx, booktabs}
\usepackage{adjustbox}
\def\BibTeX{{\rm B\kern-.05em{\sc i\kern-.025em b}\kern-.08em
    T\kern-.1667em\lower.7ex\hbox{E}\kern-.125emX}}

\begin{document}

\title{Communication-Efficient and Personalized Federated Lottery Ticket Learning\\
%\thanks{This work is supported by Institute of Information \& communications Technology Planning \& Evaluation (IITP) grant funded by the Korea government (MSIT), No.2018-0-00170, Korea-EU 5G joint project: Virtual Presence in Moving Objects through 5G (PriMO-5G). Online: https://primo-5g.eu.}
}

\author{\IEEEauthorblockN{Sejin Seo$^*$, Seung-Woo Ko$^\dagger$, Jihong Park$^\ddagger$, Seong-Lyun Kim$^*$, and Mehdi Bennis$^\S$}\\
\IEEEauthorblockA{$^*$\textit{School of EEE}, \textit{Yonsei University}, Seoul, Korea, email: \{sjseo, slkim\}@ramo.yonsei.ac.kr}
\IEEEauthorblockA{$^\dagger$\textit{Division of EEE, Korea Maritime and Ocean University}, Busan, Korea, email: swko@kmou.ac.kr}
\IEEEauthorblockA{$^\ddagger$\textit{School of Info. Tech., Deakin University}, Geelong, Australia, email: jihong.park@\{deakin.edu.au,\;gist.ac.kr\} }
\IEEEauthorblockA{$^\S$\textit{Centre for Wireless Comm.}, \textit{University of Oulu}, Oulu, Finland, email: mehdi.bennis@oulu.fi}
}

\maketitle

\begin{abstract}
The lottery ticket hypothesis (LTH) claims that a deep neural network (i.e., ground network) contains a number of subnetworks (i.e., winning tickets), each of which exhibiting identically accurate inference capability as that of the ground network. Federated learning (FL) has recently been applied in LotteryFL to discover such winning tickets in a distributed way, showing higher accuracy multi-task learning than Vanilla FL. Nonetheless, LotteryFL relies on unicast transmission on the downlink, and ignores mitigating stragglers, questioning scalability. Motivated by this, in this article we propose a personalized and communication-efficient federated lottery ticket learning algorithm, coined \textsf{CELL}, which exploits downlink broadcast for communication efficiency. Furthermore, it utilizes a novel user grouping method, thereby alternating between FL and lottery learning to mitigate stragglers. Numerical simulations validate that \textsf{CELL} achieves up to 3.6\% higher personalized task classification accuracy with 4.3x smaller total communication cost until convergence under the CIFAR-10 dataset.
\end{abstract}


\section{Introduction}
%\tred{[JH: Use notations and terminologies consistently throughout the paper (e.g., LTL, LotFL, LotteryFL)] Noted}

\begin{figure*}[t]
  \centering
  \subfigure[Baseline 1: Vanilla FL (\textsf{FedAvg}).]{\includegraphics[height=7cm]{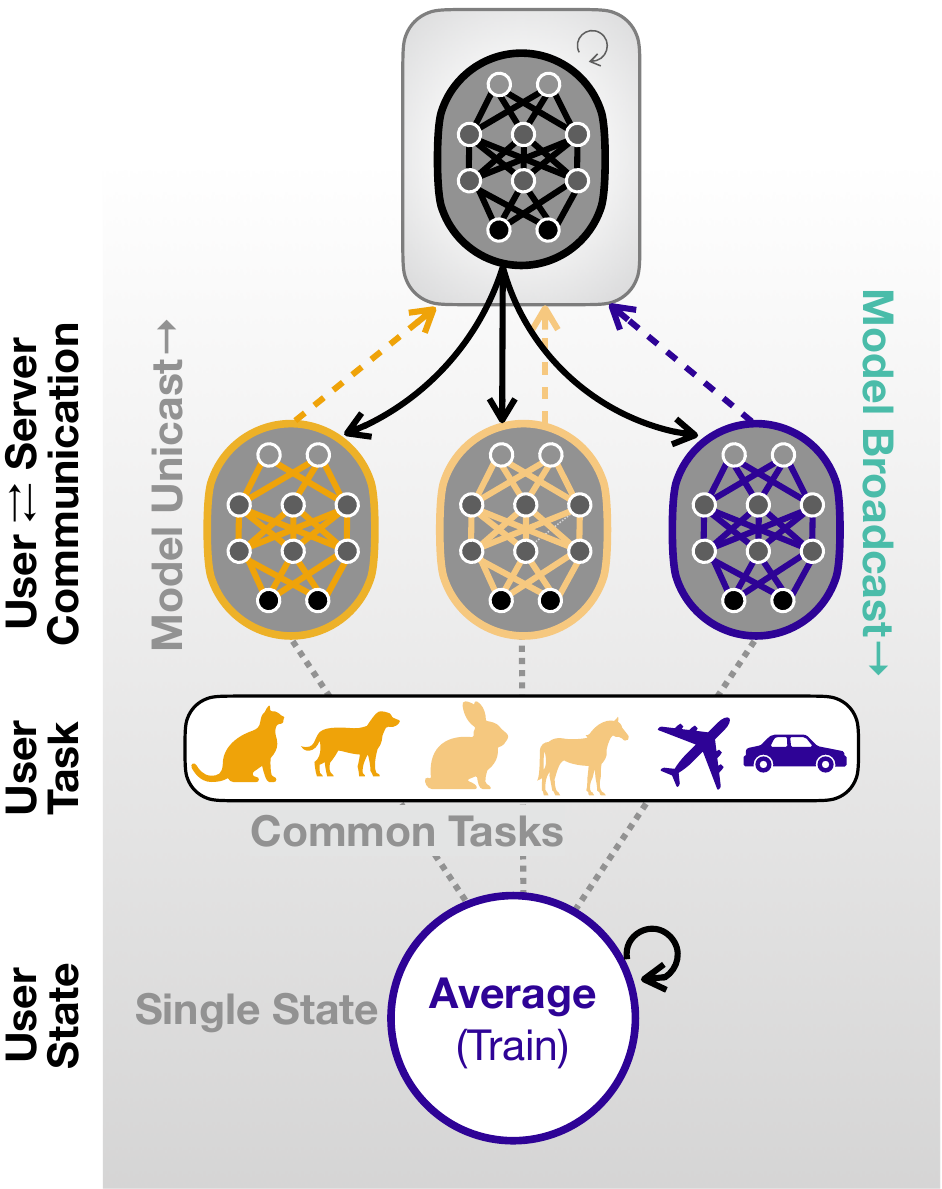}} \hspace{10pt}
  \subfigure[Baseline 2: \textsf{LotteryFL}.]{\includegraphics[height=7cm]{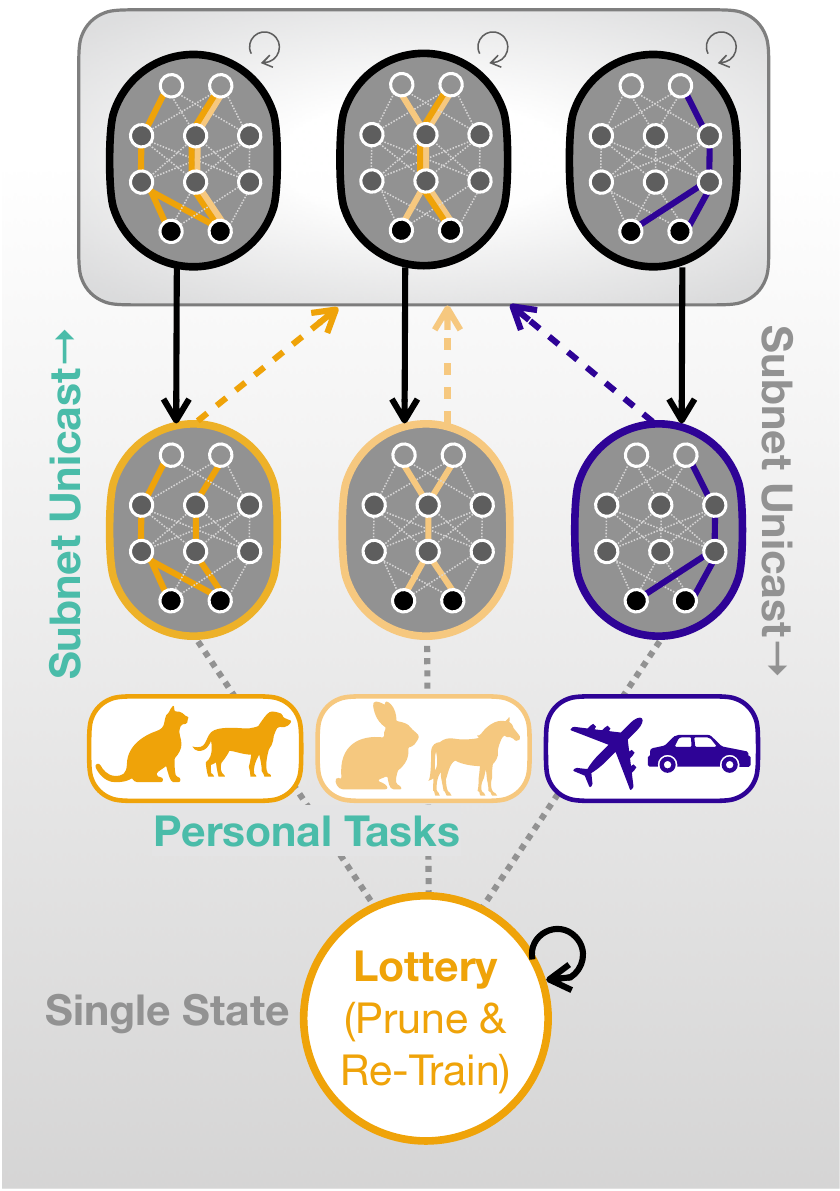}} \hspace{10pt}
  \subfigure[Proposed: \textsf{CELL}.]{\includegraphics[height=7cm]{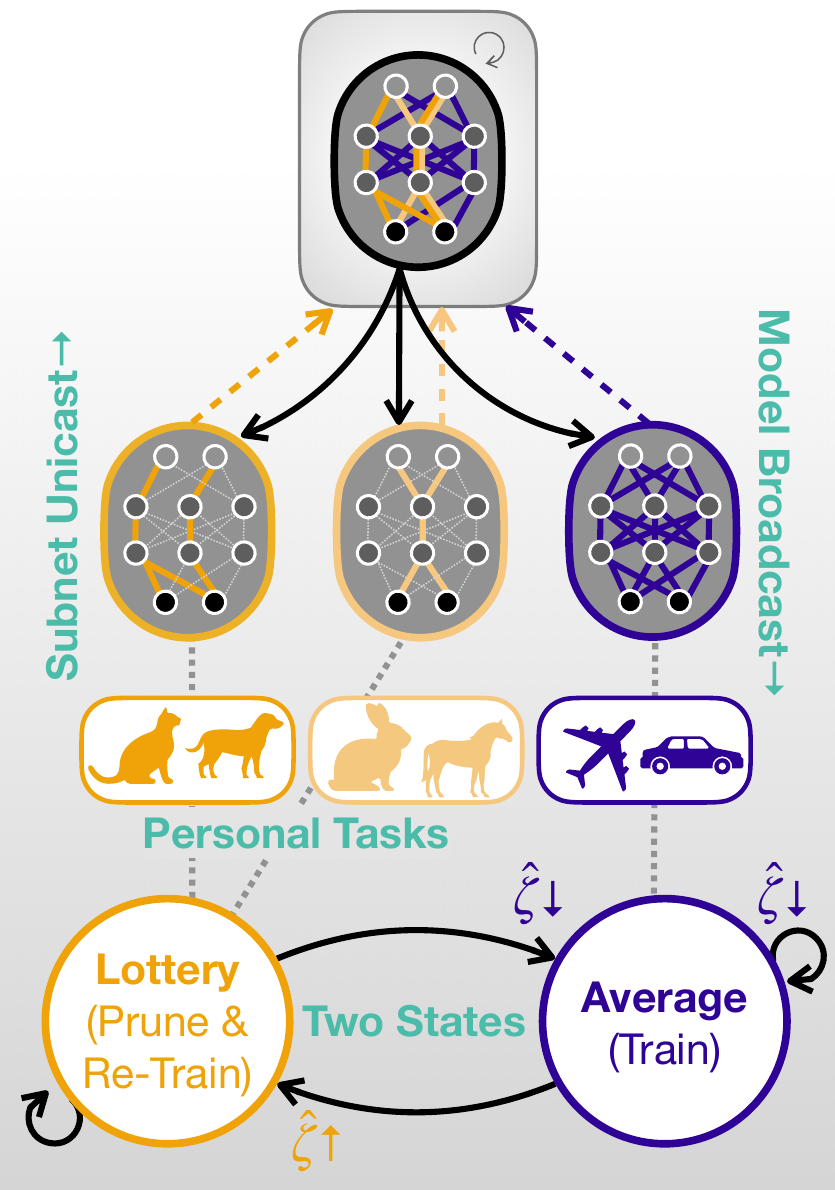}} 
  \caption{Schematic Illustrations of: (a) Vanilla federated learning (FL or \textsf{FedAvg}) with common tasks across users, (b) lottery ticket FL (\textsf{LotteryFL} with different personal tasks, and (c) communication-efficient and personalized federated lottery ticket learning (\textsf{CELL}) with personal tasks and $2$ alternating user states based on a pre-defined accuracy threshold $\hat{\zeta}$.}
\end{figure*}

Everyone is different, yet we can still learn together \cite{LC}. This captures the essence of federated learning (FL) in which multiple users collaboratively train their local models using different personal datasets \cite{FA}. In doing so, as depicted in Fig.~1a, Vanilla FL builds a single global model for all users by averaging their local models, while postulating that all users share a single common task. However, what if each user wants to learn something different than others? Imagine such a personalized learning scenario like watching Youtube or listening to Clubhouse livestreams. All the participants there experience the same global interactions between the speakers and listeners, but are interested in gaining locally tailored knowledge for each individual. Vanilla FL often fails to cope with these multiple personal tasks \cite{FML}. A n\"aive solution would be to make each user additionally train the final global model after convergence using its local data. This is however not effective particularly for deep neural networks (DNNs) with a huge number of parameters that are hardly tunable using the small-sized personal dataset of each user \cite{park2018wireless,park2020cml}

Alternatively, analogous to sifting through global knowledge and pulling out locally tailored knowledge, each user can prune the received global model parameters so as to yield a personalized local model. This pruned model enables fine-turning using the small-sized local dataset. The key to success is not to lose accuracy after pruning. The lottery ticket hypothesis (LTH) advocates its feasibility, positing that a dense DNN (i.e., ground network) contains a large number of subnetworks (i.e., winning lottery tickets) whose inference capability is as accurate as that of the ground network \cite{LTH}. Iterative pruning is the first algorithm proving the LTH by sequentially iterating: (re-)training, pruning, and re-initializing the remaining model parameters to their initial values at the ground network. The resultant final subnetwork can be trained using only a few samples without compromising accuracy compared to the ground network. 

Integrating FL into iterative pruning, as visualized in Fig.~1b, lottery federated learning (LotteryFL) enables distributed discovery of winning tickets by iterating the following operations \cite{LFL}: each user (i) prunes its local model (or subnetwork), (ii) re-trains the local model after the re-initialization, (iii) uploads the resultant local subnetwork to a server averaging the subnetworks, and finally (iv) downloads the averaged parameters replacing those of the local subnetwork. While effective under stable connectivity with unlimited bandwidth, the communication efficiency and personalization capability become questionable under intermittent wireless connectivity with limited bandwidth. To be precise, (iv) entails unicasting different subnetworks to multiple users, which is not scalable under limited bandwidth. What is more, once a user's subnetwork becomes less correlated with the others' at a certain communication round, the user can no longer contribute to other subnetworks nor be influenced by them. In other words, LotteryFL only cares about fast learners with high temporary task correlations, and ignore straggling learners who may bloom later, failing to reach the full potential.

Spurred by the aforementioned opportunities and limitations, we propose a novel communication-efficient and personalized federated lottery ticket learning (\textsf{CELL}). As illustrated in Fig.~1c, the novelty compared to LotteryFL is twofold. \begin{itemize}
    \item[1)] \textbf{Downlink Model Broadcast}: Instead of unicasting different subnetworks in the downlink, \textsf{CELL} broadcasts the global model superimposing all the current local subnetworks as well as previously accumulated parameters, in the hope of finding better subnetworks in the next~round.
    
    \item[2)] \textbf{Lottery and Federated Learning Alternation}: After downloading the global model, the users whose accuracies exceed a target threshold $\hat{\zeta}$ perform lottery learning, i.e., (i)-(ii) in LotteryFL. The remaining straggling users train the global model without pruning, i.e., Vanilla FL, followed by decreasing $\hat{\zeta}$. The last step is for promoting them to participate in lottery learning in the next round.
\end{itemize}

Under the multi-task classification for $3$ randomly chosen labels per worker of the CIFAR-10 dataset originally with $10$ labels, numerical simulations corroborate that \textsf{CELL} achieves up to $3.6\%$ higher accuracy thanks to 2). Furthermore, the total accumulated uplink-downlink communication payload size is $4.3$x smaller than LotteryFL owing to 1).

\vspace{5pt}\noindent\textbf{Related Works.}\quad
In essence, FL aims to build a one-fits-all model by exchanging and averaging multiple local models \cite{FA,park2018wireless,park2020cml}. The effectiveness of federation is thus compromised or even negated when multiple users have distinct personal tasks \cite{FML, CFL, LFL, HFL}. To account for such multitask learning issues, the solutions are broadly categorized into two directions. One way is explicitly clustering the users based on their task similarity calculated, for instance, using the global information on the dataset statistics and loss functions \cite{CFL, FML, HFL}. This is not always feasible as local datasets and loss functions are often private in FL, not to mention its incurring additional computing and communication overhead. We instead follow the other way in which the users are clustered by measuring accuracies and imposing binary masks on their local models, leading to model pruning~\cite{LFL, Sup}. 

% We take the latter approach for communication and computation efficiency, because the former approach assumes additional information other than the representation and introduces computation loads for determining the similarity.

% - \emph{Federated Multitask Learning}\cite{FML, CFL, LFL, HFL}: Both the multitask learning and federated learning literature are extensive, but they mostly focus on the single objective setting. Despite their effectiveness for a single objective setting, the federated learning and multitask learning framework cannot fully address the personalized learning problem where each user has a distinct objective \cite{CFL}, and it is not communication efficient \cite{LFL}. For instance, if we use a na\"{i}ve approach that adapts the federated averaging algorithm, the commonality of user representations are learned successfully \cite{FA}, but the heterogeneity caused by the personal tasks cannot be learned \cite{CFL}. 

Traditionally, model pruning using binary masks has been studied mostly in the context of model compression \cite{EPP}. Recently, it has been revisited thanks to the LTH stating that the masks of a dense DNN, i.e., winning subnetworks, may contain sufficient inference capabilities \cite{LTH}. The follow-up works such as \cite{SM} have empirically shown that winning subnetworks even outperform their ground network in terms of accuracy and sensitivity to initialization. Standing on these prior works, in this paper we aim to search for winning binary masks for multitask FL.

\section{Problem Definition: Searching for Winning Tickets in Federated Multitask Learning}
 
A user in \hspace{0.5mm}$\mathcal{U} = \{u_1, \cdots, u_K\}$, indexed by $k$, has a personal task with a distinct objective, e.g. learning how to classify a dog from a cat. It has exclusive training data $\mathbf{T}_k$ that could be used for finding a representation, i.e. DNN, $f_{\mathbf{w}_k} (\mathbf{x}_k)$ that fits the objective. However, when data is insufficient, the representation does not generalize to the entirety of its task. Thus, the user must cooperate with others to obtain its representation. The challenge is that most users have different objectives from itself.

To address the challenge of learning separate objectives for each user, inspired by the multitask learning formulation in \cite{FML}, we formulate the problem of our interest as follows:

\vspace{-5pt}\small\begin{align}  \label{P1}
\tag{P1} \min_{\mathbf{W}, \mathbf{M}} \quad & \! \sum_{k=1}^{K}\!{\ell_k(\mathbf{w}_k\! \circ \!\mathbf{m}_k)} \!+\! \lambda_1 \underbrace{tr\!\left(\!\mathbf{W}\! \circ\! \mathbf{M} \mathbf{\Omega} \! \left(\!\mathbf{W}\! \circ\! \mathbf{M}\!\right)^\mathrm{T}\!\right)}_{\textrm{multitask regularizer}}\! +\!\lambda_2\!\underbrace{\left\|\mathbf{W}\! \circ\! \mathbf{M}\right\|_2^2}_{\textrm{LTH reguarlizer}}, 
% \\
% \textrm{where} \quad & |\mathbf{w}_k \circ \mathbf{m}_k| \leq |\mathbf{w}_k|, \forall k, \nonumber 
\end{align}\normalsize
where $\mathbf{W}:= \left[\mathbf{\mathbf{w}}_1,\dots,\mathbf{\mathbf{w}}_K\right] \in \mathbb{R}^{d\times K}$ is the matrix whose $k$-th column is the model weights of the $k$-th user,  $\mathbf{M}:= \left[\mathbf{m}_1,\dots,\mathbf{m}_K\right] \in \mathbb{R}^{d\times K}$ is the matrix whose $k$-th column is the binary mask $\mathbf{m}_k \in \mathbb{R}^{d}$ of the $k$-th user, yielding $|\mathbf{w}_k \circ \mathbf{m}_k| \leq |\mathbf{w}_k|$. The term $\mathbf{\Omega}$ is the correlation matrix for the tasks, and $\lambda_1, \lambda_2 > 0$ are regularization constants, and $\hat{\gamma}\in(0,1)$ is the target pruning rate. The operation $\circ $ is the element-wise product of matrices, and the superscript $\mathrm{T}$ denotes the matrix transpose. The multitask regularizer follows from \cite{FML}, penalizing highly correlated tasks, thereby promoting to learn more diverse personal tasks. The LTH regularizer, on the other hand, penalizes large models, promoting more model pruning.

% the overlap between less correlated tasks, while the LTH regularizer promotes 
% and the $L_2$ regularizer penalizes large weights and dense models. 

% \tred{SJ:}

% The regularizer $\mathcal{R}$ is defined as follows:

% \begin{equation} \label{S1}
% \begin{aligned}
% \mathcal{R}(\mathbf{W}\!,\! \mathbf{M}\!,\! \mathbf{\Omega})\!=\!\lambda_1 tr\!\left(\!\left(\mathbf{W}\! \circ\! \mathbf{M}\right) \!\mathbf{\Omega}^{-1} \! \left(\mathbf{W}\! \circ\! \mathbf{M}\right)^T\!\right)\! +\!\lambda_2\!\left\|\mathbf{W}\! \circ\! \mathbf{M}\right\|_2^2\\
% \end{aligned}
% \end{equation}

\section{Communication Efficient and Personalized Federated Lottery Ticket Learning}

%Problem \eqref{P1} requires each user to find a winning ticket that is sufficiently dense, i.e. dimension of $\mathbf{m}_k \circ \mathbf{w}_k$ is closer to the dimension of $\mathbf{w}_k$, to help in the federated averaging process, and sufficiently sparse, i.e. dimension of $\mathbf{m}_k \circ \mathbf{w}_k$ closer to $0$, to prevent its model from exerting too much influence on others, especially the ones that are less correlated to itself. 

The works in the federated multitask learning literature either assume \emph{a priori} knowledge regarding the task correlations $\mathbf{\Omega}$ \cite{MTL}, assume that the objective functions are biconvex \cite{FML}, or calculate the correlation directly by using additional information \cite{CFL}. Problem \eqref{P1} becomes more challenging as the assumptions above get lifted. Also, the LTH regularizer in \eqref{P1} further complicates the problem, because it impacts the task correlation as the rounds progress. Also, lottery ticket related researches \cite{LTH, Rigl, Sup} mostly focus on finding a single winning ticket for a single task, which cannot be directly translated to the multitask problem. 

%So the questions at hand are, how do we find such winning tickets in a distributed environment where everyone wants to learn something different? And, how do we find them efficiently? 
To utilize lottery tickets in a distributed environment, Li et al. proposes a possible direction in \cite{LFL} by combining the procedures of federated learning \cite{FA} and lottery ticket hypothesis \cite{LTH}, calling it lottery federated learning (\textsf{LotteryFL}). As mentioned in the introduction, \textsf{LotteryFL} runs with the following operations: each user (i) prunes its local model (or subnetwork), (ii) re-trains the local model after the re-initialization, (iii) uploads the resultant local subnetwork to a server averaging the subnetworks, and finally (iv) downloads the averaged parameters replacing those of the local subnetwork. Despite the novelty of the idea, the work still lacks the specified design for heterogeneous learning objectives.

In a heterogeneous task environment, we need to understand that some tasks have loss landscapes with much lower optimum points, difficult convergence characteristics, or poor transfer learning property. This becomes evident when we compare users by evaluating the validation accuracy $\zeta\in[0,1]$ of a global aggregate $\mathbf{w}_g$ w.r.t. the user's validation data $\mathbf{V}_k$. A fast learner's high validation accuracy, i.e. $\zeta \geq \hat{\zeta}$, either signifies that the current global aggregate contains sufficient knowledge regarding the user's personal task, or indicates that the personal task is easy. On the other hand, a straggler's low validation accuracy, i.e. $\zeta < \hat{\zeta}$, indicates the opposite. Furthermore, Even for the same task, the loss landscape changes as the round progresses, as the aggregated global model changes and user models get sparser. 

Therefore, if we enforce the same validation threshold $\hat{\zeta}\in[0,1]$ to all users at all rounds like \textsf{LotteryFL}, stragglers will not be able to start searching for its winning ticket. This not only hurts the test accuracy of each user, but also harms the communication efficiency of the algorithm, i.e. LTH regularizer $\left\|\mathbf{W}\! \circ\! \mathbf{M}\right\|_2^2$ getting stuck at a certain level. Thus, we propose the communication efficient lottery learning (\textsf{CELL}) which uses an adaptive threshold for each user, according to its history of failure regarding the validation test $\zeta \gtreqless \hat{\zeta}$. 

\textsf{CELL}'s strength lies in straggler control dictated by the following points. Firstly, each time a user becomes a straggler, the user is granted a higher chance to enter the winning ticket search by lowering its $\hat{\zeta}$. Secondly, a straggler not allowed to enter the winning ticket search at round $t$ skips pruning and re-initialization, i.e. $\mathbf{w}_k^t = \mathbf{w}^0_g \circ \mathbf{m}_k^t$. This lets the straggler exert more influence to the global model $\mathbf{w}_g^t$ at the round, because its model $\mathbf{w}_k^t$ is denser and more overfitted than the fast learners. 

By subsequently pruning the model of users in the order of their task difficulty (from easiest to hardest), fairness is ensured for all stragglers to exert their influence eventually. After a user is given a chance to search its winning ticket, $\hat{\zeta}$ is restored to the default, e.g. $0.5$, to differentiate the users again, according to their task difficulty. These procedures regarding $\hat{\zeta}$ serve as a moderate clustering of users according to their task difficulty, because users with similar difficulty share similar masks. The moderate clustering effect reduces the interference between the heterogeneous features between tasks, while maintaining the aggregation for the common features. Lastly, unlike \textsf{LotteryFL}, by broadcasting the full global model $\mathbf{w}_g^t$, the users are given a fresh attempt at finding a new winning ticket, i.e. $\mathbf{m}_k^t \neq \mathbf{m}_k^{t+1}$, at every round. 

Furthermore, \textsf{CELL} is communication efficient regarding both uplink and downlink transmissions. The uplink efficiency is achieved by letting all users, including fast learners and stragglers, to prune their models at the right time, and the downlink efficiency is achieved by broadcasting the global model to the users

The detailed explanation for the procedure is written below as a supplement for the pseudocode.

\begin{table}[t]
  \centering
  \label{tab:alg1}
  \begin{tabular}{c l}
    \toprule
    \multicolumn{2}{l}{\textbf{Algorithm 1}: Communication-Efficient Lottery Learning (\textsf{CELL})}\\
    \multicolumn{2}{l}{User $u_k \in \mathcal{U}$ has validation data $\mathbf{V}_k$ and train data $\mathbf{T}_k$}\\
    \midrule
    1:& \hspace{-4mm} \textbf{Server executes:}\\
    2:& \hspace{-2mm} initialize $\mathbf{w}^0_g$\\
    3:& \hspace{-2mm} \textbf{for} communication round $t = 1,\cdots$ \textbf{do}\\
    4:& \hspace{0mm} \vline \hspace{1mm} $p$ $\xleftarrow{}$ max$(C \cdot |\mathcal{U}|, 1)$ \hspace{2mm}$\triangleright$ $C$ ratio of users participate  \\
    5:& \hspace{0mm} \vline \hspace{1mm} $S_t$ $\xleftarrow{}$ \{randomly sampled $p$ users\}\\
    6:& \hspace{0mm} \vline \hspace{1mm} \textbf{for} each user $k \in S_t$ \textbf{in parallel do} \\
    7:& \hspace{0mm} \vline \hspace{3mm} \vline \hspace{1mm} $\mathbf{w}^{t+1}_k$ $\xleftarrow{}$ \textsf{CELL}($k$, $\mathbf{w}^{t}_g$, $\mathbf{w}^0_g$) \\
    8:& \hspace{0mm} \vline \hspace{1mm} $\mathbf{w}^{t+1}_g \xleftarrow{} \sum_{k\in S_t}{\frac{n_k}{n} \mathbf{w}^{t+1}_k}$  \hspace{2mm}$\triangleright$ do federated averaging\\
    9:& \hspace{-4mm} \textbf{\textsf{CELL}($k$, $\mathbf{w}^t_g$, $\mathbf{w}^0_g$):} \\
    10:& \hspace{-2mm} $\zeta \xleftarrow{}$ validation accuracy of $\mathbf{w}^{t}_g$ w.r.t. $\mathbf{V}_k$ \\
    11:& \hspace{-2mm} $\gamma^t_k \xleftarrow{}$ previously pruned amount of user $k$ \\
    12:& \hspace{-2mm} \textbf{if} $\gamma^t_k < \hat{\gamma}$: \hspace{4mm}$\triangleright$ compare with target pruning rate\\
    13:& \hspace{0mm} \vline \hspace{1mm} $\gamma^{t+1}_k \xleftarrow{}$ min($\gamma^t_k + \gamma$, $\hat{\gamma}$) \\
    14:& \hspace{0mm} \vline \hspace{1mm} \textbf{if} $\zeta > \hat{\zeta}$:  \hspace{2mm}$\triangleright$ compare with validation acc. threshold\\
    15:& \hspace{0mm} \vline \hspace{3mm} \vline \hspace{1mm} prune $\gamma^{t+1}_k$ weights from $\mathbf{w}^{t}_k$\\
    16:& \hspace{0mm} \vline \hspace{3mm} \vline \hspace{1mm} reinitialize $\mathbf{w}^{t}_k$ with $\mathbf{w}^0_g$\\
    17:& \hspace{0mm} \vline \hspace{1mm} \textbf{else}\\
    18:& \hspace{0mm} \vline \hspace{3mm} \vline \hspace{1mm} decrease $\hat{\zeta}$ to $\alpha*\hat{\zeta}$\\
    19:& \hspace{0mm} \vline \hspace{1mm} $\mathbf{w}^{t+1}_k \xleftarrow[]{}$ \textbf{train for} $E$ epochs with $\mathbf{T}_k$ \\
    20:& \hspace{-2mm} \textbf{else}:\\
    21:& \hspace{0mm} \vline \hspace{1mm} prune $\hat{\gamma}$ from $\mathbf{w}^{t}_k$\\
    22:& \hspace{0mm} \vline \hspace{1mm} $\mathbf{w}^{t+1}_k \xleftarrow[]{}$ \textbf{train for} $E$ epochs with $\mathbf{T}_k$ \\
    23:& \hspace{-2mm} \textbf{return} $\mathbf{w}^{t+1}_k$ to server\\
  \bottomrule
\end{tabular}
\end{table}

\subsection{Step 1. Global Model Validation}

At each round $t\in\{1,\cdots\}$ a user indexed by $k$ evaluates the global model $\mathbf{w}^t_g$ with the local validation data $\mathbf{V}_k$, which sufficiently describes the task that the user has. If the validation accuracy $\zeta$ exceeds a pre-defined validation threshold $\hat{\zeta}\in[0,1]$, proceed to Step 2-1. lottery ticket learning; otherwise, enter Step 2-2. federated learning with validation control.

\subsection{Step 2-1. Lottery Ticket Learning}

To find the winning tickets, following the key principles of LTH, we adhere to the procedures as shown below.

\begin{itemize}
    \item[1)] \textbf{Magnitude Based Pruning}: Prune the network w.r.t. the weight magnitude, i.e. $\|w\|$  $\forall w \in \mathbf{w}_k$, to gain a subnetwork that has only $\gamma^t_k \in (0,1]$ of the original weights. One step of pruning increments the pruning rate by $\gamma$, and the pruning rate stops increasing after the target pruning rate $\hat{\gamma}$ is reached. 
    
    \item[2)] \textbf{Re-Initialization and Re-Training}: Re-initialize the subnetwork with the initial weights, i.e. $\mathbf{w}_k^t = \mathbf{w}^0_g \circ \mathbf{m}_k^t$, then retrain the subnetwork for $E$ epochs.
    
    \item[3)] \textbf{Uploading}: Upload the trained subnetwork, i.e. $\mathbf{w}_k^{t+1} \circ \mathbf{m}_k^t$, to the server.
    
\end{itemize}

\subsection{Step 2-2. FL with Validation Threshold Control}

\begin{itemize}
    \item[1)] \textbf{Training}: For $E$ local epochs, train the dense network, i.e. $\mathbf{w_k} = \mathbf{w_k} - \eta \nabla \ell(\mathbf{w_k};b)$, where $\eta$ is the learning rate and $b$ is the minibatch.
    
    \item[2)] \textbf{Uploading}: Upload the dense model, i.e. $\mathbf{w}_k^{t+1}$, to the server. 
    
    \item[3)] \textbf{Validation Threshold Ramping}: For each failed attempt at entering Step 2-1, the validation threshold $\hat{\zeta}$ is multiplied by $\alpha\in[0,1)$.
\end{itemize}

\subsection{Step 3. Aggregation and Broadcast}
Server aggregates the subnetworks received from the users at Step 1 and the full models received from the users at Step 2, i.e. $\mathbf{w}^{t+1}_g = \sum_{k\in S_t}{\frac{n_k}{n} \mathbf{w}^{t+1}_k}$, where $n_k$ is the number of local data and $n=\sum n_k$. The denser models exert more influence than the sparser models, and the sparser models focus their knowledge aggregation at the location of their masks. To speed up model sparsification while taking in larger networks, the total pruning amount is incremented at each winning ticket discovery, i.e. $\gamma^{t+1}_k = \min(\gamma^t_k + \gamma$, $\hat{\gamma})$, until the target pruning rate $\hat{\gamma}$ is achieved.

\begin{figure*}[t]
  \centering
  \includegraphics[width=18cm]{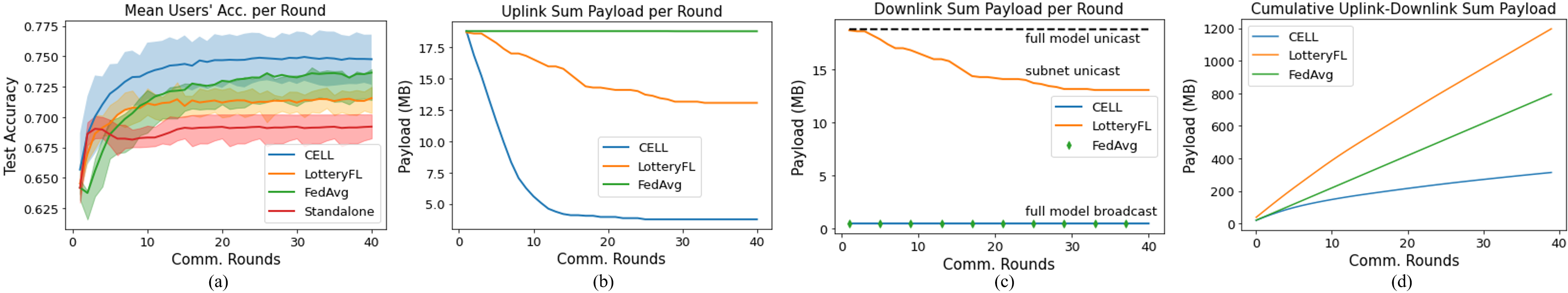}
  \caption{Test accuracy and communication cost comparison for the federated multitask learning problem in a synchronous setting, $C=1, n_k=100$: (a) Users' test accuracies are averaged at each communications round; (b) uplink payload (MB) per round; (c) downlink payload (MB) per round; and (d) cumulative communication cost (MB) shows the total uplink and downlink payloads of the system.} \label{Perf}
\end{figure*}

\section{Performance Evaluation}

%#\subsection{Baseline}
%\begin{itemize}

%\item Standalone: Each user trains its own model by using only its local training dataset until convergence.
%\item FedAvg \cite{FA}: At each communication round, each user participating in the federated learning process trains its model by using its local training dataset for $E$ epochs, then uploads the model to the server. The server aggregates the models received at the round by using a weighted sum w.r.t. the number of data each user used for training, forming the global model for the round. Then, the server returns the global model to the users participating in the next round.  
%\item LotteryFL \cite{LFL}: At each communication round, each user validates its model w.r.t. its validation data. If validation accuracy exceeds certain threshold, it is pruned with a predefined amount. The pruned model is re-initialized as in \cite{LTH}, then retrained with local training dataset. users upload their models to the server, and the aggregation step is identical to FedAvg. At the next round, the server returns only the aggregated model that is pruned by the current mask of the corresponding user.
%\end{itemize}

\subsection{Simulation Settings}
\begin{itemize}
\item Model: Each user uses a CNN model with $2$ convolution layers and $3$ fully connected layers like the CNN model in \cite{FA}. This is a variation of LeNet-$5$\cite{Lenet} that replaces a convolution layer with fully connected layer.
\item Dataset: CIFAR-$10$ dataset is used.
%, which consists of $50{,}000$ train samples and $10{,}000$ test samples, each representing a $32\times32 (\times3$ RGB channels) image sample among 10 classes. 
The train dataset is sorted by label and partitioned into shards of $n_k$ samples. 
%Most shards contain only one label, but some could contain $2$ or more, depending on $n_k$. With this setting, when users sample $3$ batches of $n_k$ data, most user holds data from two or more labels, thus having a distinct task for itself. 
The test dataset for a user consists the entire test data for each label that the user sampled for training. $n_k=100$ and $B=3$ are used. 
\item Training: Each user is trained for $E$ local epochs. Minibatch stochastic gradient descent method, i.e. $\mathbf{w} = \mathbf{w} - \eta \nabla \ell(\mathbf{w};b)$, where $\eta$ is the learning rate and $b$ is the minibatch of size $B$. $E=10$ and $B=32$ are used for performance evaluation.
\item Pruning: Magnitude (L1-norm) based one-shot unstructured pruning is applied over weight parameters only.%, because parameters like bias and batch-normalization are negligible in parameter count but may affect performance negatively. 
For the tests, $\gamma=0.2$ and $\hat{\gamma}=0.8$ are chosen.
\item Time-Variance: The number of participants in each round can be tuned to see the effect of time-variance. The tests use participation ratios $C \in \{0, 0.1, 0.2, 0.5, 1\}$. For $C = 0$, we consider one participant at each round.
\item Baseline: Standalone, \textsf{FedAvg} \cite{FA}, and \textsf{LotteryFL} \cite{LFL} are compared with \textsf{CELL}.
\end{itemize}

\subsection{Test Accuracy}

From Fig. \ref{Perf}(a), \textsf{CELL} outperforms the other models' personalized task accuracy on average. The performance increase is evident from the early rounds, due to \textsf{CELL}'s aggressive pruning strategy. 

As shown in Table \ref{TestAcc}, the evidence becomes stronger for the asynchronous setting ($C<1$), as \textsf{FedAvg}'s performance plummets, when either local data is insufficient or participation ratio is low. Roughly speaking, \textsf{FedAvg} starts performing once it has more than $3{,}000$ samples used for a round ($C=0.5, n_k=50$), while the lottery based algorithms start performing with as little as $240$ samples ($C=0.1, n_k=20$). The solid performance of the lottery based algorithms in highly asynchronous settings serves as an evidence that the users with easy tasks exert too much influence on the global aggregate at when the total data is insufficient. In addition, \textsf{CELL} outperforms \textsf{LotteryFL} with consistent margin. This performance gain is an evidence that \textsf{CELL}'s achievement of fairness, i.e. subsequently giving harder tasks opportunities exert more influence, helps all users, including fast learners and stragglers, to learn better.

\subsection{Communication Efficiency}

As shown in Fig. \ref{Perf}(b)-(d), \textsf{CELL} shows significant improvement in communication efficiency. Especially, \textsf{CELL}'s uplink (UL) efficiency in Fig. \ref{Perf}(b) is promising for cellular communications based learning systems, given the highly unbalanced nature of uplink (UL) and downlink (DL) performance of commercial 5G \cite{5GP}, i.e. UL having much lower capacity. As shown in Table \ref{CommCost} under different settings, \textsf{CELL}'s UL efficiency is resilient against diverse tasks, because it prunes users' models in even the harshest environment, consisting of highly heterogeneous tasks w.r.t. their difficulties and the level of correlation between them.
In addition, \textsf{CELL}'s broadcasting architecture allows its downlink payload to be independent from the number of participating users. This makes \textsf{CELL} scalable for systems with large number of users. These results are surprising, because higher level of model sparsity often degrades performance \cite{LTH}.

\begin{table}[]
\centering
\caption{Test Accuracy after $40$ Communication Rounds} \label{TestAcc}
\begin{adjustbox}{width=\columnwidth}
\begin{tabular}{cccccccccc}
\toprule
\multicolumn{10}{c}{CIFAR-$10$: Personalized Test Accuracy (\%)}\\
 & \multicolumn{3}{c}{\textsf{\color{ForestGreen}FedAvg}} & \multicolumn{3}{c}{\textsf{\color{orange}LotteryFL}} & \multicolumn{3}{c}{\textsf{\color{blue}CELL}} \\
 & \multicolumn{3}{c}{\textbf{---------}$n_k$\textbf{---------}} & \multicolumn{3}{c}{\textbf{---------}$n_k$\textbf{---------}} & \multicolumn{3}{c}{\textbf{---------}$n_k$\textbf{---------}} \\
$C$& \multicolumn{1}{c}{$20$} & \multicolumn{1}{c}{$50$} & \multicolumn{1}{c}{$100$} & \multicolumn{1}{c}{$20$} & \multicolumn{1}{c}{$50$} & \multicolumn{1}{c}{$100$} & \multicolumn{1}{c}{$20$} & \multicolumn{1}{c}{$50$} & \multicolumn{1}{c}{$100$} \\ \hline
$0.0$&$28.9$&$26.3$&$28.3$&$\mathbf{40.4}$&$41.6$&$45.7$&$40.3$&$\mathbf{41.8}$&$\mathbf{45.9}$\\
$0.1$&$39.6$&$41.8$&$44.5$&$62.6$&$70.4$&$70.8$&$\mathbf{64.1}$&$\mathbf{72.5}$&$\mathbf{71.3}$\\
$0.2$&$39.5$&$42.1$&$61.0$&$60.9$&$71.6$&$71.0$&$\mathbf{65.8}$&$\mathbf{73.9}$&$\mathbf{72.2}$\\
$0.5$&$38.6$&$69.7$&$70.8$&$62.7$&$72.5$&$71.7$&$\mathbf{65.7}$&$\mathbf{74.3}$&$\mathbf{73.0}$\\
$1.0$&$60.2$&$72.8$&$\color{ForestGreen}73.3$&$64.6$&$\color{orange}72.6$&$70.8$&$\mathbf{65.3}$&$\mathbf{\color{blue}74.6}$&$\mathbf{74.4}$\\ 
\bottomrule
\end{tabular}
\end{adjustbox}
\end{table}

\begin{table}[]
\centering
\caption{Communication Cost after $40$ Communication Rounds} \label{CommCost}
\begin{adjustbox}{width=\columnwidth}
\begin{tabular}{cccccccc}
\toprule
\multicolumn{8}{c}{CIFAR-$10$: Total Comm. Payload (MB)}\\
 & \multicolumn{1}{c}{\textsf{\color{ForestGreen}FedAvg}} & \multicolumn{3}{c}{\textsf{\color{orange}LotteryFL}} & \multicolumn{3}{c}{\textsf{\color{blue}CELL}} \\
 & \multicolumn{1}{c}{\textbf{---}$n_k$\textbf{---}} & \multicolumn{3}{c}{\textbf{------------}$n_k$\textbf{------------}} & \multicolumn{3}{c}{\textbf{------------}$n_k$\textbf{------------}} \\
$C$& \multicolumn{1}{c}{$20, 50, 100$} & \multicolumn{1}{c}{$20$} & \multicolumn{1}{c}{$50$} & \multicolumn{1}{c}{$100$} & \multicolumn{1}{c}{$20$} & \multicolumn{1}{c}{$50$} & \multicolumn{1}{c}{$100$} \\ \hline
$0.0$ & $58.9$  & $36.7$ & $37.2$ & $\mathbf{34.8}$ & $\mathbf{35.1}$ & $\mathbf{35.0}$ & $35.5$ \\
$0.1$ & $115.4$ & $146.7$ & $147.6$ & $147.4$ & $\mathbf{81.2}$ & $\mathbf{84.3}$ & $\mathbf{81.5}$ \\
$0.2$ & $190.8$ & $280.9$ & $277.8$ & $276.4$& $\mathbf{126.9}$ & $\mathbf{127.2}$ & $\mathbf{120.2}$ \\
$0.5$ & $\color{ForestGreen}417.0$ & $656.9$ & $\color{orange}609.1$ & $603.9$ & $\mathbf{206.6}$ & $\mathbf{204.0}$ & $\mathbf{\color{blue}196.2}$ \\
$1.0$ & $794.0$ & $1312.0$ & $1286.0$ & $1195.9$ & $\mathbf{298.2}$ & $\mathbf{305.1}$ & $\mathbf{313.2}$\\ 
\bottomrule
\end{tabular}
\end{adjustbox}
\end{table}

\section{Conclusion and Future Work}

In this work, we propose \textsf{CELL} to address federated multitask learning in a communication efficient manner. \textsf{CELL} achieves remarkable reduction of communication cost while ensuring the personal task performance, by exploiting the heterogeneous difficulty of users' personal tasks to adjust the threshold for starting the lottery learning process. The ability of masks to filter information in a federated learning environment is intriguing yet underexplored. Though we considered only basic unstructured pruning combined with the basic lottery ticket hypothesis procedure, one can extend this to other pruning methods and lottery learning process that could aggregate the commonalities and differentiate the heterogeneity better. Also, even higher communication efficiency and privacy could be achieved if the masks can become primary agents for carrying information, rather than a supplementary filter for the information.

\section*{Acknowledgment}
%\begin{acks}
\small
This work was supported in part by Institute of Information \& communications Technology Planning \& Evaluation (IITP) grant funded by the Korea government (MSIT), No.2018-0-00170, Korea-EU 5G joint project: Virtual Presence in Moving Objects through 5G (PriMO-5G, Online: https://primo-5g.eu), and in part by IITP grant No. 2014-3-00077.

\normalsize

%\end{acks}


\begin{thebibliography}{00}

\bibitem{LC} J. Roschelle, ``Learning by collaborating: Convergent conceptual change,'' \emph{J. Learn. Sci.}, Vol. 2, No.3, pp. 235-276, 1992.

\bibitem{FA} H. B. McMahan et al., ``Communication-efficient learning of deep networks from decentralized data,'' in \emph{Proc. Int. Conf. on Artif. Intell. Statist. (AISTATS)}, 2017, FL, USA.

\bibitem{FML} V. Smith et al., ``Federated multi-task learning,'' in Proc. \emph{Adv. Neural Inf. Process. Syst.}, 2017, pp. 4424–4434.

\bibitem{park2018wireless}
J.~Park, S.~Samarakoon, M.~Bennis, and M.~Debbah, ``Wireless network
  intelligence at the edge,'' {\em Proceedings of the IEEE}, vol.~107,
  pp.~2204--2239, October 2019.
  
  \bibitem{park2020cml}
J.~Park, S.~Samarakoon, A.~Elgabli, J. Kim, M.~Bennis, S.-L, Kim, and M.~Debbah, ``Communication-Efficient and Distributed Learning Over Wireless Networks: Principles and Applications,'' {\em to appear in Proceedings of the IEEE}.


\bibitem{LFL} A. Li et al., ``LotteryFL: Personalized and communication-efficient federated learning with lottery ticket hypothesis on non-IID datasets,'' arXiv:2008.03371, 2020, [Online] Available: http://arxiv.org/abs/2008.03371. 

\bibitem{CFL} F. Sattler, K.-R. M\"{u}ller, and W. Samek, ``Clustered federated learning: Model-agnostic distributed multitask optimization under privacy constraints,'' \emph{IEEE Trans. Neural Netw. Learn. Syst.}, Early Access, pp. 1-13, Aug. 2020.

\bibitem{HFL} A. Ghosh et al., ``Robust federated learning in a heterogeneous environment,'' 2019, arXiv:1906.06629. [Online]. Available: http://arxiv.org/abs/1906.06629


\bibitem{Sup} M. Wortsman et al., ``Supermasks in superposition,'' in Proc. \emph{Annu. Conf. Neural Info. Process. Syst. (NeurIPS)}, 2020, Vancouver, Canada.


\bibitem{EPP} H. Wang et al., ``Emerging paradigms of neural network pruning,'' arXiv:2103.06460, 2021, [Online] Available: https://arxiv.org/abs/2103.06460.

\bibitem{LTH} J. Frankle and M. Carbin, ``The lottery ticket hypothesis: Finding sparse, trainable neural networks,'' in \emph{Proc. Int. Conf. Learn. Represent. (ICLR)}, p. 42, Apr. 2018, Vancouver, Canada.

\bibitem{SM} H. Zhou et al., ``Deconstructing lottery tickets: Zeros, signs, and the supermask,'' in Proc. \emph{Annu. Conf. Neural Info. Process. Syst. (NeurIPS)}, 2019, Vancouver, Canada.

\bibitem{MTL} A. Argyriou, T. Evgeniou, and M. Pontil, ``Multi-task feature learning,'' in Proc. \emph{Neural Info. Process.
Syst. (NIPS)}, 2007.

\bibitem{Rigl} U. Evci et al., ``Rigging the lottery: Making all tickets winners,'' in Proc. \emph{Int. Conf. Mach. Learn.}, July 2020, Virtual.

\bibitem{Lenet} Y. LeCun, ``LeNet-5, convolutional neural networks,'' 2015, [Online] Available: http://yann.lecun.com/exdb/lenet.


\bibitem{5GP} S. Seo, S. Kim, and S.-L. Kim, ``A public safety framework for immersive aerial monitoring through 5G commercial network,'' in Proc. \emph{IEEE Wireless Commun. Netw. Conf. Workshops (WCNC)}, pp. 1-6, Apr. 6-9, 2020, Seoul, Korea.






\end{thebibliography}
\end{document}